\documentclass[lettersize,journal]{IEEEtran}
\usepackage{amsmath,amsfonts}
\usepackage{algorithm}
\usepackage{array}
\usepackage{textcomp}
\usepackage{stfloats}
\usepackage{url}
\usepackage{verbatim}
\usepackage{graphicx}
\usepackage{cite}
\usepackage{algorithm}
\usepackage{algpseudocode}

\usepackage{multirow}
\usepackage{tikz}
\usepackage{booktabs} 
\usepackage{multicol}
\usepackage{subfigure}
\usepackage{threeparttable}
\usepackage{mathrsfs}
\usepackage{makecell}

\usepackage[table]{xcolor}       
\definecolor{mygray}{gray}{.9}   

\makeatletter
\newcommand{\thickhline}{%
\noalign {\ifnum 0=`}\fi \hrule height 1pt
\futurelet \reserved@a \@xhline
}
\makeatother
\newcommand{\pub}[1]{\color{gray}{\tiny{[{#1}]}}}

\usepackage[table]{xcolor}       
\definecolor{mygray}{gray}{.9}   
\definecolor{myblue}{RGB}{220,235,250}
\definecolor{myyellow}{RGB}{220,235,250} 
\PassOptionsToPackage{numbers}{natbib}
\definecolor{url}{RGB}{0,73,147}
\definecolor{mypink}{HTML}{bc4749}

\usepackage{xspace}
\usepackage{bbding}
\makeatletter
\DeclareRobustCommand\onedot{\futurelet\@let@token\@onedot}
\def\@onedot{\ifx\@let@token.\else.\null\fi\xspace}

\def\ie{\emph{i.e}\onedot}

\makeatother

\usepackage{amsmath,amsfonts,bm}

\def\eqref#1{equation~\ref{#1}}

\def\1{\bm{1}}

\DeclareMathAlphabet{\mathsfit}{\encodingdefault}{\sfdefault}{m}{sl}
\SetMathAlphabet{\mathsfit}{bold}{\encodingdefault}{\sfdefault}{bx}{n}

\begin{document}

\title{A New Multi-Domain Benchmark for Micro-Action Recognition and Detection}

\author{Yanbin Hao*,~\IEEEmembership{Member,~IEEE}, Pengyu Liu*, Xing Wei, Xun Yang, Dan Guo,~\IEEEmembership{Senior Member,~IEEE}, Meng Wang$^{\dagger}$,~\IEEEmembership{Fellow,~IEEE}
\thanks{Yanbin Hao, Pengyu Liu, Xing Wei, Dan Guo and Meng Wang are with the School of Computer Science and Information Engineering, Hefei University of Technology, Hefei, China (e-mail: haoyanbin@hotmail.com,  lpynow@gmail.com, xing.owners@gmail.com, guodan@hfut.edu.cn, eric.mengwang@gmail.com).}
\thanks{Xun Yang is with the School of Information Science and Technology, University of Science and Technology of China, Hefei, China (e-mail: xyang21@ustc.edu.cn).}
\thanks{*: Co-first author.}
\thanks{$^{\dagger}$: Corresponding author.}
}


\markboth{Journal of \LaTeX\ Class Files,~Vol.~14, No.~8, August~2021}%
{Shell \MakeLowercase{\textit{et al.}}: A Sample Article Using IEEEtran.cls for IEEE Journals}


\maketitle

\begin{abstract}
Micro-actions are short-duration, low-amplitude subtle body movements at the whole-body level that can reveal latent intentions, involuntary reactions, and fine-grained affective changes. Our previous MA-52 benchmark has provided an important foundation for micro-action recognition, but it remains limited in scale, scene diversity, task coverage, and evaluation protocols. To advance micro-action analysis toward more realistic and comprehensive settings, we introduce \textbf{MMA-82}, a large-scale multi-domain extension of MA-52. MMA-82 expands the label space from 52 to 82 fine-grained micro-action categories and covers four distinct domains, including laboratory interviews, street interviews, psychiatric patient interviews, and emotion-rich television videos, resulting in 79,574 annotated instances from 454 subjects. Built upon MMA-82, we establish two core tasks: \textbf{Micro-Action Recognition} and \textbf{Multi-label Micro-Action Detection}. For recognition, we further define in-domain and cross-domain protocols, including few-shot and zero-shot settings, to evaluate model robustness, transferability, and generalization. Extensive experiments show that current methods still struggle with realistic micro-action understanding, especially under domain shift, long-tailed category distributions, and complex temporal localization. Beyond benchmarking, we investigate the relationship between micro-actions and emotion, showing that micro-actions are strongly associated with emotional states and provide complementary cues to facial micro-expressions for improved emotion recognition. These results demonstrate that MMA-82 serves as a comprehensive and challenging benchmark for realistic micro-action analysis and a valuable resource for human-centered AI.  MMA-82 is available at \url{https://lpynow.github.io/MMA-82-AIM/}.
\end{abstract}


\vspace{-0.5cm}
\section{Introduction}
\begin{figure*}[t]
\centering
\vspace{-2em}
\subfigure[Comparison between existing micro-gesture/action datasets and our MMA-82. ``MG'' is the combination of iMiGUE and SMG datasets.]{
    \includegraphics[width=\linewidth]{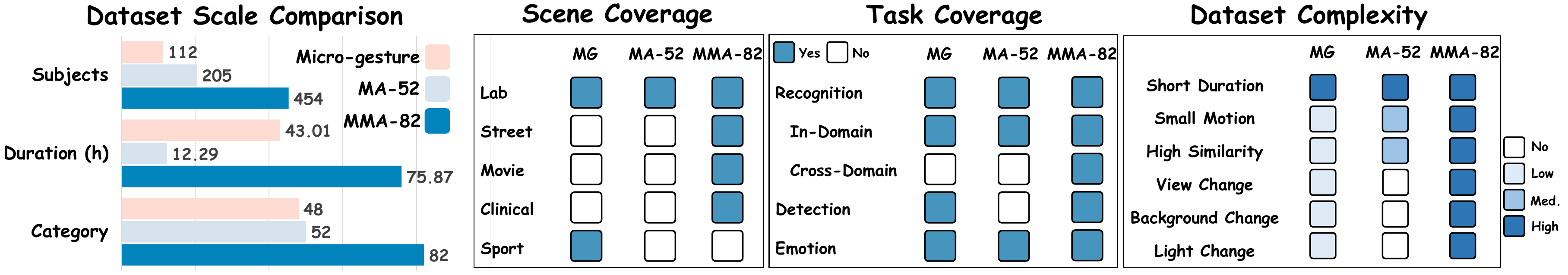}
    \label{fig:compare_main_figure}
}
\hfill
\subfigure[Representative micro-action video samples from four sources in our MMA-82.]{
    \includegraphics[width=\linewidth]{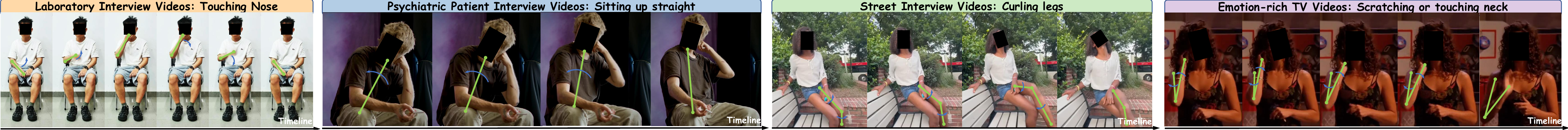}
    \label{fig:video_frame}
}
\vspace{-1em}
\caption{Overview of the proposed MMA-82 benchmark.}
\vspace{-2em}
\label{fig:overview}
\end{figure*}

\IEEEPARstart{H}{uman} action recognition in computer vision has long been dominated by the study of macro-actions, namely voluntary and goal-directed activities characterized by large motion amplitudes and relatively long temporal durations, such as running, playing, or object manipulation. This research line has steered the development of large-scale benchmarks, such as UCF101~\cite{soomro2012ucf101}, Kinetics~\cite{carreira2017quo}, and Something-Something~\cite{goyal2017something}, as well as powerful recognition models, including SlowFast~\cite{feichtenhofer2019slowfast}, Video Swin~\cite{liu2022video} and GPT4Ego~\cite{dai2024gpt4ego}. However, macro-actions mainly capture coarse and explicit aspects of human behavior, while subtle yet behaviorally informative motion patterns, namely \textbf{micro-actions}, remain largely underexplored.

\textbf{Micro-actions refer to short-duration, low-amplitude subtle body movements at the whole-body level} that may convey important cues about human intention, emotional state, and spontaneous response. Analogous to facial micro-expressions, which reveal concealed affective states through fleeting facial changes~\cite{mao2025facial}, micro-actions reflect fine-grained motion dynamics that may disclose latent intentions, unconscious reactions, and incipient behavioral tendencies. Such signals are potentially valuable for a wide range of applications, including human-computer interaction, security screening, healthcare monitoring, and sports analytics.

Recognizing micro-actions is inherently challenging, which largely explains their scarcity in current research. First, micro-actions are extremely short in duration, often lasting only a few frames, making them difficult to capture reliably and easy to overlook during annotation and recognition. Second, their motion amplitudes are typically very small, so the relevant signals can be easily obscured by background clutter, camera noise, illumination variations, natural body fluctuations, and other irrelevant motions. Third, collecting micro-action data at scale is particularly challenging, since these behaviors are often spontaneous, weakly controlled, and highly context-dependent, making them difficult to elicit, record, segment, and annotate consistently. Nevertheless, studying micro-actions is of fundamental importance: humans can naturally infer intention and internal state from subtle body movements, and enabling machines to acquire such fine-grained perceptual ability is crucial for advancing embodied intelligence and human-machine interaction.

Our previous MA-52~\cite{guo2024benchmarking} is the pioneering benchmark specifically introduced for micro-action recognition and marked an important early milestone in this emerging research area. It has already attracted growing community attention: to date, we have organized three editions of the \emph{Micro-Action Analysis Grand Challenge} in ACM Multimedia 2024\&2025\&2026, involving more than 100 participating teams, and benchmark performance has improved substantially over time. Building upon MA-52, a series of follow-up studies, including MMAD~\cite{li2025mmad}, MA-Bench~\cite{li2026bench}, Motion Matters~\cite{gu2025motion}, and PCAN~\cite{li2025prototypical}, have further advanced this field by developing stronger models and extending the task scope from recognition to more challenging settings such as detection and QA-oriented tasks. Nevertheless, despite its pioneering role and growing impact, MA-52 still has limited capability to support large-scale and systematic study. Its data are predominantly collected in controlled laboratory-style environments with fixed viewpoints, stable illumination, clean backgrounds, and constrained body postures, which are far removed from the complexity and unpredictability of real-world scenarios. Moreover, its scale and annotation granularity remain limited in terms of videos, subjects, scenes, and action categories, while its task and protocol coverage is still relatively narrow, making it difficult to systematically assess model performance under realistic conditions such as cross-domain transfer, low-resource learning, and more challenging recognition settings.

Accordingly, our work is centered around the following four research questions (\textbf{RQs}): 
\begin{itemize}
    \item \textbf{RQ1:} \emph{\textbf{How can we build a realistic micro-action benchmark beyond controlled laboratory environments?}}
    \item \textbf{RQ2:} \emph{\textbf{How can micro-action understanding be evaluated more comprehensively in realistic settings?}}
    \item \textbf{RQ3:} \emph{\textbf{Are micro-actions associated with emotional states in videos?}}
    \item \textbf{RQ4:} \emph{\textbf{Do micro-actions provide complementary affective cues beyond facial micro-expressions?}}
\end{itemize}
 These questions jointly target the central bottlenecks of current micro-action research, including realistic data collection, comprehensive evaluation, and the broader behavioral significance of subtle whole-body-level motion patterns.

To answer these questions, we introduce \textbf{MMA-82}, a comprehensive multi-domain benchmark for micro-action understanding. MMA-82 is designed to extend and improve MA-52 in a structured and one-to-one manner. \textbf{\emph{On the data side}}, it expands the source domain from a single laboratory interview setting to four distinct domains, including laboratory interview videos, street interview videos, psychiatric patient interview videos, and emotion-rich television videos, thereby providing substantially richer variations in scene context, viewpoints, illumination, background clutter, recording conditions, and subject diversity. \textbf{\emph{On the benchmark side}}, MMA-82 not only enlarges the label space from 52 to 82 categories, but also supports both \emph{Micro-Action Recognition} and \emph{Multi-label Micro-Action Detection}, together with \emph{in-domain}, \emph{cross-domain}, \emph{few-shot}, and \emph{zero-shot} evaluation settings. Beyond benchmark construction, MMA-82 further enables us to investigate the relationship between micro-actions and emotions, and to examine whether subtle body movements provide complementary affective cues beyond facial micro-expressions. 

Our main contributions are summarized as follows:
\begin{itemize}
\item We present \textbf{MMA-82}, a comprehensive multi-domain benchmark for micro-action understanding. Compared with our prior version MA-52, MMA-82 substantially improves realism, scale, category richness, scene diversity, and overall complexity, providing a more rigorous testbed for subtle whole-body-level human action understanding.

\item We establish two core benchmark tasks, namely Micro-Action Recognition and Multi-label Micro-Action Detection, together with in-domain, cross-domain, few-shot, and zero-shot evaluation protocols. These settings enable systematic assessment of model robustness, transferability, and generalization under realistic conditions.

\item We conduct extensive experiments and analyses on MMA-82. The results reveal the intrinsic difficulty of realistic micro-action understanding, especially under domain shift, long-tailed category distributions, and challenging temporal localization scenarios.

\item We investigate the behavioral significance of micro-actions and show that they are strongly associated with emotional states. We further demonstrate that micro-actions provide complementary affective cues to micro-expressions for emotion recognition.

\end{itemize}

\begin{table*}[]
\centering
\caption{Comparative overview of relevant datasets. ``MA'', ``MG'' and ``ME'' denote micro-action/gesture/expression, respectively.}
\vspace{-0.2cm}
\begin{threeparttable}
\resizebox{1.0\linewidth}{!}{
\setlength{\tabcolsep}{2.5pt}
\begin{tabular}{p{0.3cm} || l l |l||c|c|c|c|c|c|c|c|c || l}
\thickhline
\rowcolor{mygray}
& \multicolumn{2}{c||}{\textbf{Dataset}}   & \multicolumn{1}{c||}{\textbf{Venue}} & \textbf{Resolution} & \textbf{Labels} & \textbf{Instances} & \textbf{Duration} (h) & \textbf{Frames} & \textbf{Subjects} & \textbf{Emotion Labels} & \multicolumn{1}{c|}{\textbf{Scene}} & \multicolumn{1}{c||}{\textbf{Body Area}} & \multicolumn{1}{c}{\textbf{Task}} \\ \thickhline\hline


\multirow{2}{*}{%
  \rotatebox[origin=c]{90}{%
    MG
  }%
}

&
\multicolumn{2}{l||}{iMiGUE~\cite{liu2021imigue}}
& \pub{CVPR'21}   
& 1280×720   
& 32              
& 18,499          
& 34.87           
& 3.14M           
& 72              
& 2               
& Real            
& Up-body         
& Action Recognition \\

&
\multicolumn{2}{l||}{SMG~\cite{chen2023smg}}
& \pub{IJCV'23}  
& 1920×1080
& 17
& 3,712       
& 8.14    
& 0.82M   
& 40    
& 2 
& Lab     
& Whole body
& Action Detection \\ \hline\hline

\multirow{6}{*}{%
  \rotatebox[origin=c]{90}{%
    ME
  }%
}

&
\multicolumn{2}{l||}{CAER~\cite{lee2019context}}
& \pub{ICCV'19}
& -
& - 
& 20,484
& 12.82
& 1.11M
& -
& 7
& TV videos
& Whole body
& Emotion Recognition\\

&
\multicolumn{2}{l||}{SAMM~\cite{davison2016samm}}
& \pub{TAC'16}
& 2040×1088
& -
& 159
& -
& -
& 32
& 8
& Lab
& Face
& Emotion Recognition\\

&
\multicolumn{2}{l||}{SMIC~\cite{li2013spontaneous}}
& \pub{FG'13}
& 640×480
& -
& 164
& -
& -
& 16
& 3
& Lab
& Face 
& Emotion Recognition\\

&
\multicolumn{2}{l||}{CASME~\cite{yan2013casme}}
& \pub{FG'13}
& 640×480
& -
& 195
& -
& -
& 19
& 7
& Lab
& Face
& Emotion Recognition\\ 

& 
\multicolumn{2}{l||}{CASME II~\cite{qu2017cas}}
& \pub{TAC'17}
& 640×480
& -
& 247
& -
& -
& 26
& 5
& Lab
& Face
& Emotion Recognition\\ 

&
\multicolumn{2}{l||}{CAS(ME)$^3$~\cite{li2022cas}}
& \pub{TPAMI'22}
& 1280×720
& -
& 1,030
& 8
& 8M
& 216
& 7
& Lab
& Face
& Emotion Recognition\\ \hline\hline

\multirow{5}{*}{%
  \rotatebox[origin=c]{90}{%
    MA
  }%
}

&
\multicolumn{2}{l||}{MA-52~\cite{guo2024benchmarking}}
& \pub{TCSVT'24}
& 1080×1296
& 52
& 22,422          
& 12.29   
& 1.33M  
& 205    
& 5 
& Lab
& Whole body    
& Action Recognition\\ 

&
\multicolumn{2}{l||}{MMA-52~\cite{li2025mmad}}
& \pub{ICCV'25}
& 1080×1296
& 52
& 19,782         
& 18.67 
& 2.01M  
& 203   
& -
& Lab
& Whole body    
& Action Detection\\ \cline{2-14}

&
\multicolumn{2}{>{\columncolor{myyellow}}l||}{MMA-82 (\textbf{ours})}
& \cellcolor{myyellow}-     
& \cellcolor{myyellow}-         
& \cellcolor{myyellow}\textbf{82}         
& \cellcolor{myyellow}\textbf{79,574}         
& \cellcolor{myyellow}\textbf{75.87}       
& \cellcolor{myyellow}\textbf{8.05M}       
& \cellcolor{myyellow}\textbf{454}                        
& \cellcolor{myyellow}-             
& \cellcolor{myyellow}Mix      
& \cellcolor{myyellow}Whole body  
& \cellcolor{myyellow}\textbf{Both}\\ 

&
\multicolumn{2}{>{\columncolor{myyellow}}l||}{--MMA-82-Rec}
& \cellcolor{myyellow}-     
& \cellcolor{myyellow}Mix         
& \cellcolor{myyellow}\textbf{82}         
& \cellcolor{myyellow}\textbf{39,816}         
& \cellcolor{myyellow}\textbf{28.94}       
& \cellcolor{myyellow}\textbf{2.99M}       
& \cellcolor{myyellow}\textbf{454}                        
& \cellcolor{myyellow}8             
& \cellcolor{myyellow}Mix      
& \cellcolor{myyellow}Whole body        
& \cellcolor{myyellow}Action Recognition\\ 

&
\multicolumn{2}{>{\columncolor{myyellow}}l||}{--MMA-82-Det}
& \cellcolor{myyellow}-    
& \cellcolor{myyellow}Mix           
& \cellcolor{myyellow}\textbf{82}         
& \cellcolor{myyellow}\textbf{39,758}         
& \cellcolor{myyellow}\textbf{46.93}         
& \cellcolor{myyellow}\textbf{5.06M}
& \cellcolor{myyellow}\textbf{434}          
& \cellcolor{myyellow}- 
& \cellcolor{myyellow}Mix      
& \cellcolor{myyellow}Whole body           
& \cellcolor{myyellow}Action Detection\\ \cline{2-14}

\thickhline
\end{tabular}
}
\end{threeparttable}
\label{tab:compare_dataset_main}
\vspace{-0.5cm}
\end{table*}

\vspace{-0.2cm}
\section{Related Work}

In this section, we compare MMA-82 with existing human subtle motion datasets, including micro-expression, micro-gesture, and micro-action benchmarks. We also briefly review several representative fine-grained action datasets to position our work within the broader landscape of video action understanding. Fig.~\ref{fig:overview} and Table~\ref{tab:compare_dataset_main} summarize the differences between MMA-82 and representative datasets in terms of scale, scene coverage, task coverage, and dataset complexity.

In video action understanding, several datasets focus on relatively fine-grained human actions, including Something-Something~\cite{goyal2017something}, NTU RGB+D 60~\cite{shahroudy2016ntu}, NTU RGB+D 120~\cite{liu2019ntu}, EPIC-KITCHENS~\cite{9084270},  Ego4D~\cite{grauman2022ego4d} and SAV~\cite{tan2025towards}. Compared with more coarse-grained action datasets such as UCF101~\cite{soomro2012ucf101} and Kinetics~\cite{carreira2017quo}, these datasets place greater emphasis on fine-grained human-object interactions, procedural differences, and temporal reasoning. However, they are still mainly designed for explicit, long-duration, and semantically salient actions, rather than brief, subtle, and weakly controlled body motions.

In the broader area of subtle human behavior analysis, many existing datasets focus on facial expressions or upper-body gestures. CASME II~\cite{yan2014casme} and SMIC~\cite{li2013spontaneous} are two representative micro-expression datasets. Micro-expressions are subtle and involuntary facial muscle movements that occur within an extremely short duration and often reveal genuine emotional states. Micro-gesture datasets, such as iMiGUE~\cite{liu2021imigue} and SMG~\cite{chen2023smg}, focus on subtle upper-body and hand movements that may reflect hidden emotions, suppressed intentions, or stress-related responses. These datasets demonstrate the practical value of subtle body cues for affective and stress-related analysis, but they mainly capture local facial, upper-body, or hand movements rather than subtle whole-body motions.

MA-52~\cite{guo2024benchmarking} was the first benchmark dedicated to whole-body micro-actions, collected in a laboratory interview setting, with MMA-52~\cite{li2025mmad} further developed on top of it for the detection task. Building upon MA-52, our MMA-82 advances micro-action research toward a larger-scale, more diverse, and more realistic benchmark. Compared with MA-52, MMA-82 expands the number of action categories from 52 to 82, increases the numbers of videos and subjects, and broadens the data sources from a single laboratory interview scenario to four distinct domains. Moreover, MMA-82 establishes a more comprehensive evaluation framework, including two core benchmark tasks, micro-action recognition and multi-label micro-action detection, together with in-domain and cross-domain protocols, where the latter is further divided into few-shot and zero-shot settings. As a result, MMA-82 provides not only a substantial extension in scale and diversity but also a more comprehensive benchmark for advancing micro-action recognition and detection under realistic conditions.

\vspace{-0.2cm}
\section{The MMA-82 Benchmark}
MMA-82 is constructed through a multi-stage pipeline designed to extend prior micro-action resources toward richer category coverage, higher-quality temporal annotations, and more generalizable samples. As an extension of MA-52, MMA-82 aims to support more realistic and systematic micro-action understanding across diverse scenarios. Specifically, the construction process consists of the following stages: \textbf{(A)} defining and refining the extended label system, \textbf{(B)} collecting large-scale data from multiple sources, and \textbf{(C)} training annotators and establishing annotation guidelines. Based on this pipeline, we further build two benchmark datasets, namely MMA-82-Rec for micro-action recognition and MMA-82-Det for multi-label micro-action detection. The detailed annotations and benchmark settings of MMA-82-Rec and MMA-82-Det will be introduced in the next section. Through its multi-domain data collection, extended 82-category taxonomy, and multi-stage annotation pipeline, MMA-82 directly answers \textbf{RQ1} by providing a realistic, large-scale, and richly annotated benchmark beyond controlled laboratory environments.

\vspace{-0.4cm}
\subsection{Label Definition and Extension}

\begin{figure*}[t]
\centering
\includegraphics[width=\linewidth]{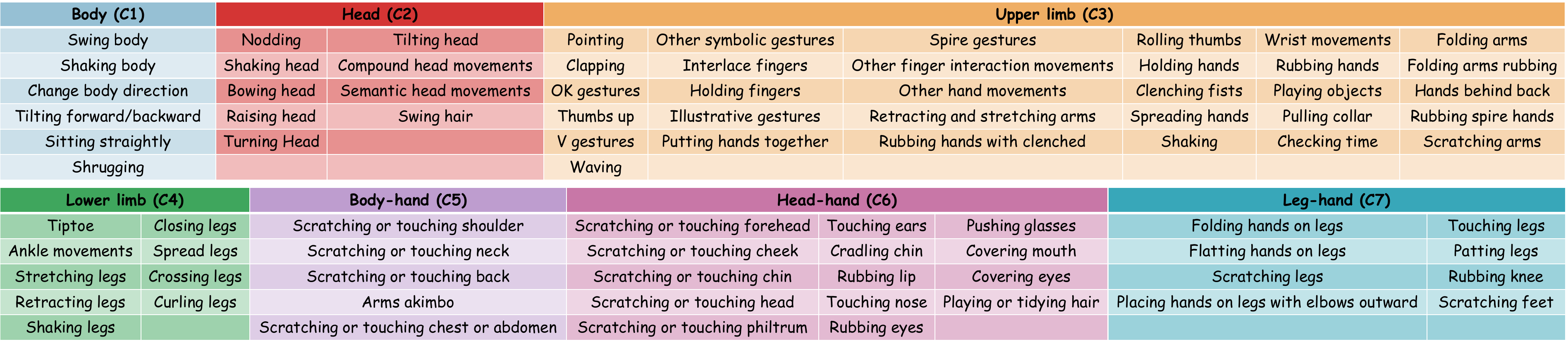}
\caption{MMA-82 comprises 7 \texttt{Body-level} and 82 \texttt{Action-level} micro-actions, covering the majority of common micro-action categories.}
\label{fig:label_define}
\vspace{-0.5cm}
\end{figure*}

To build a more comprehensive micro-action taxonomy, we expand the 52 categories of MA-52 to 82 categories. The extended label set places greater emphasis on fine-grained whole-body movement patterns and is organized into seven coarse-grained groups: C1: Body, C2: Head, C3: Upper Limb, C4: Lower Limb, C5: Body-Hand, C6: Head-Hand, and C7: Leg-Hand, as shown in Fig.~\ref{fig:label_define}. Compared with the original label space, this refined taxonomy captures temporal dynamics and spatial motion patterns in greater detail, making it more suitable for describing the complexity of micro-actions in both constrained and natural scenes. For example, we additionally add `\textit{Thumbs up}', `\textit{Holding fingers}', `\textit{Folding arms}' for Upper Limb actions, and `\textit{Placing hands on legs with elbows outward}', `\textit{Rubbing knee}', `\textit{Folding hands on legs}' for Leg-Hand. These actions frequently appear in everyday nonverbal cues and carry specific meanings, which complement the micro-action structure of MA-52. This 82-category label system serves as the basis of MMA-82, supporting more precise modeling, evaluation, and analysis of micro-actions.

\vspace{-0.4cm}
\subsection{Dataset Collection}
Capturing spontaneous micro-actions is inherently challenging, because they are typically subtle in intensity, short in duration, and difficult to annotate accurately. Existing collection strategies can generally be divided into three categories. The first is \textbf{\emph{prompted performance}}, in which participants are instructed to act out predefined micro-actions, as done by iMiGUE~\cite{liu2021imigue}. This strategy is useful for collecting rare categories, but the resulting actions are often more standardized and exaggerated than naturally occurring behaviors. The second is \textbf{\emph{interview-based recording}}, where subjects are recorded during face-to-face conversations or psychological interviews, allowing more spontaneous micro-actions to emerge, as done by MA-52~\cite{guo2024benchmarking}. The third is \textbf{\emph{in-the-wild annotation}}, where micro-actions are identified and annotated from street interviews, television videos, or other uncontrolled videos, as done by SMG~\cite{chen2023smg}. 

To address the limitations of any single strategy, MMA-82 is constructed through a multi-source and multi-type data collection process. Our goal is to capture micro-actions across different environments, subjects, and recording conditions, and to improve both category diversity and real-world realism. To this end, we combine multiple collection strategies and carefully construct several complementary sub-datasets.

\textbf{Laboratory interview videos:} We reuse the interview videos from MA-52, which were collected through a specialized face-to-face psychological interview protocol based on participants' SCL-90 test results. These videos provide a valuable foundation for capturing relatively spontaneous micro-actions in a semi-controlled environment. However, some rare micro-action categories are difficult to elicit naturally and are therefore severely underrepresented in the original dataset. To alleviate this long-tailed distribution, we first analyze the category distribution of MA-52 and identify a set of rare micro-actions. We then recruit 24 additional participants and ask them to perform these rare micro-actions according to their own understanding. Although this process is guided, it helps supplement categories that are difficult to capture in sufficient quantity through natural interviews alone. At the same time, allowing participants to interpret the actions in their own way preserves individual variation in performance style and motion characteristics. This collection stage is therefore used to intentionally enrich rare categories and improve category balance in the benchmark.

\textbf{Psychiatric patient interview videos:} This subset is collected from publicly available psychiatric patient interview videos on YouTube. The videos typically consist of topic-guided self-narratives from individuals with different psychiatric disorders, covering life experiences, daily challenges, emotional fluctuations, and disorder-related cognition. Compared with laboratory-style interviews, these videos provide more realistic behavioral expressions under less constrained conditions, and thus offer a valuable resource for studying subtle micro-actions in psychiatric contexts.

\textbf{Street interview videos:} This subset is mainly collected from unscripted street interview videos on YouTube, which are highly unstructured and recorded in open, uncontrolled environments. Interview settings, camera viewpoints, and recording conditions exhibit substantial randomness and variability. The subjects often come from socially marginalized or underrepresented groups, such as homeless individuals and transgender people, and the interviews typically focus on daily survival, social exclusion, personal hardship, and past trauma. Such narratives are often accompanied by big individual differences, emotional fluctuations, and irregular speaking rhythms, which naturally induce frequent subtle body movements. At the same time, environmental noise, illumination changes, camera shake, and cluttered backgrounds make micro-action detection significantly more challenging. Therefore, this subset is particularly valuable for studying multi-label micro-action detection in realistic scenarios, and serves as an important complement to laboratory and semi-structured settings.

\textbf{Emotion-rich television videos:} This subset is derived from the CAER~\cite{lee2019context} dataset, one of the most widely used benchmarks for emotion recognition. CAER is mainly collected from television shows, documentaries, and other real-world videos, and contains rich emotional expressions in diverse natural scenes. Based on CAER, we select a subset of videos and further annotate them with fine-grained micro-actions, thereby establishing explicit links between emotional states and subtle body movements. This subset provides a useful resource for investigating how emotional changes are reflected through micro-actions and for studying the relationship between action and emotion in naturalistic settings.

\vspace{-0.4cm}
\subsection{Action Segment Annotation}

For each video, we use LabelU~\cite{he2024opendatalab} to annotate every micro-action instance, including its category label and the corresponding start and end times in the temporal sequence. To ensure annotation quality and consistency, all annotators undergo standardized and long-term professional training before participating in the annotation process. Specifically, we provide a detailed annotation manual, which describes the category definitions, classification rules, and temporal boundary criteria for each micro-action class.

To reduce noise from extremely short clips, we impose a minimum duration threshold and require each annotated action segment to last longer than 0.2 seconds. Each video is annotated by three annotators. First, two annotators independently label all action instances in the video, including the action category and the corresponding temporal boundaries. Then, a third senior annotator manually reviews the results of the first two annotators, resolves category inconsistencies, and verifies the temporal annotations. This multi-stage process helps improve both label reliability and temporal consistency.

Because temporal boundary annotation is inherently subjective, different annotators may produce slightly different start and end times for the same action instance. To reconcile these discrepancies, we adopt the annotation merging strategy proposed in EPIC-KITCHENS~\cite{damen2020epic}. Let the temporal annotation for the $i$-th action instance provided by the $j$-th annotator be denoted as $A_i^j = [t_{s,i}^{(j)}, t_{e,i}^{(j)}]$, where $t_{s,i}^{(j)}$ and $t_{e,i}^{(j)}$ represent the start and end times of the action, respectively.

To further quantify annotation agreement, we compute an agreement score based on the Intersection-over-Union (IoU) between temporal segments. For the $j$-th annotator, the inter-annotator agreement score is defined as:
\vspace{-0.8em}
\begin{equation}
    s_i^{(j)} = \frac{1}{K_a} \sum_{k=1}^{K_a} \mathrm{IoU}\!\left(A_i^{(j)}, A_i^{(k)}\right),
    \vspace{-0.5em}
\end{equation}
where $K_a$ is the number of annotators ($K_a=3$). The annotator with the maximum agreement is selected as
\vspace{-0.5em}
\begin{equation}
    \hat{j} = \arg\max_j s_i^{(j)}.
    \vspace{-0.8em}
\end{equation}
We then identify the annotation that has the highest IoU overlap with the selected annotation:
\vspace{-0.5em}
\begin{equation}
    \hat{k} = \arg\max_k \mathrm{IoU}\!\left(A_i^{(\hat{j})}, A_i^{(k)}\right).
    \vspace{-0.5em}
\end{equation}
Finally, the ground-truth temporal segment $A_i$ is determined as
\vspace{-0.8em}
\begin{equation}
    A_i =
    \begin{cases}
        \mathrm{union}\!\left(A_i^{(\hat{j})}, A_i^{(\hat{k})}\right), & \text{if } \mathrm{IoU}\!\left(A_i^{(\hat{j})}, A_i^{(\hat{k})}\right) > 0.5,\\
        A_i^{(\hat{k})}, & \text{otherwise}.
    \end{cases}
\end{equation}
As demonstrated in~\cite{damen2020epic}, this strategy aggregates the annotations with the highest agreement by taking their union, thereby preventing the resulting temporal segment from being overly tight and improving the reliability of boundary annotation.

\begin{figure*}[t]
\centering
\subfigure[Distribution of the MMA-82-Rec dataset.]{
    \includegraphics[width=0.9\linewidth]{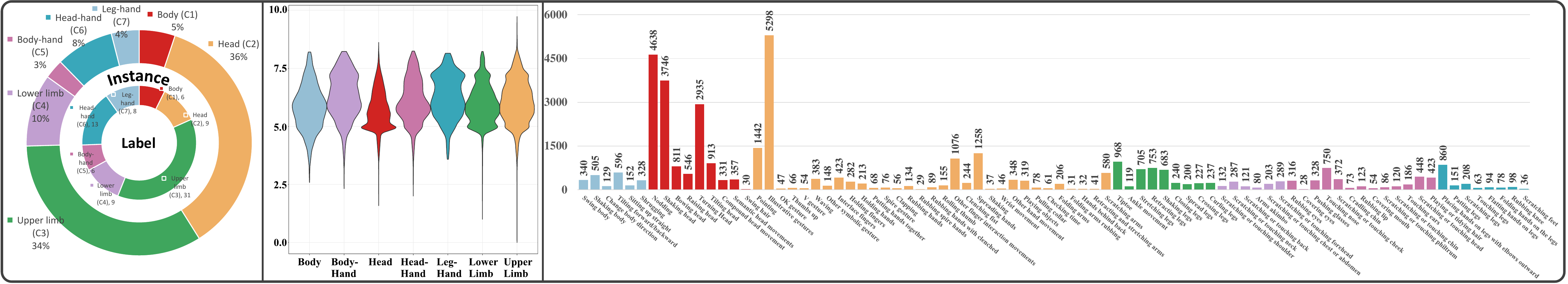}
    \label{fig:distribution_MAR}
}
\hfill
\subfigure[Distribution of the MMA-82-Det dataset.]{
    \includegraphics[width=0.9\linewidth]{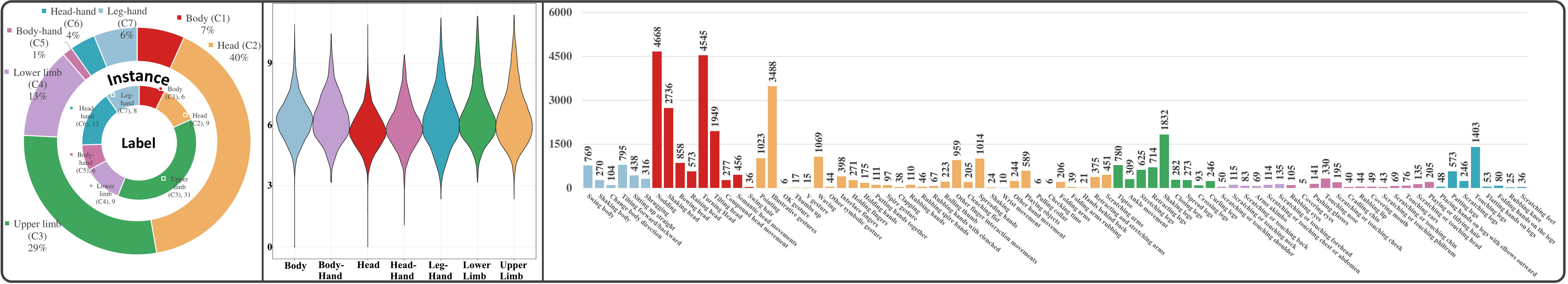}
    \label{fig:distribution_MAD}
}

\vspace{-0.2cm}
\caption{Comprehensive statistics of the MMA-82 from multiple perspectives. (a) and (b) show the statistical information of MMA-82-Rec and MMA-82-Det across different domains. The left figure of each subplot represents the \texttt{Body-Level} category distribution of action instances, with most instances concentrated in ``Head'' and ``Upper Limb.'' The middle figure shows the distribution between \texttt{Body-Level} categories and frames, exhibiting varied temporal distributions; for ease of representation, the y-axis is in $log_2^{frames}$. The right figure further provides the detailed instance distribution across all 82 fine-grained micro-action categories, showing a pronounced long-tailed pattern.}
\vspace{-0.3cm}
\label{fig:all_dataset_class_count}
\end{figure*}

\vspace{-0.2cm}
\section{Tasks and Baselines} \label{sec:benchmark}
This section answers \textbf{RQ2} by establishing recognition and detection tasks under both standard and realistic generalization settings. Specifically, we introduce two benchmark tasks that together constitute the MMA-82 benchmark suite. The first task (Sec.~\ref{sec:MAR}) is \textit{Micro-Action Recognition} (MAR), where each video clip is assigned a single micro-action label and its corresponding dataset is denoted as \textbf{MMA-82-Rec}. The second task (Sec.~\ref{sec:MAD}) is \textit{Multi-label Micro-Action Detection} (MAD), which is built upon \textbf{MMA-82-Det} dataset and aims to localize and classify multiple micro-action instances in untrimmed videos. Together, these tasks evaluate micro-action understanding at both semantic and temporal levels, encompassing category recognition and temporal localization.

\vspace{-0.4cm}
\subsection{Micro-Action Recognition}\label{sec:MAR}

\textbf{\emph{Task Definition.}} Given a video $V$,  the MAR task aims to predict the category of the target micro-action. Following our annotation hierarchy, each micro-action can be described at two levels: the \emph{body level} $\mathcal{C}_{body} \in \{1,2,\ldots, N_{body}\}$ and the \emph{action level} $\mathcal{C}_{act} \in \{1,2,\ldots,N_{act}\}$, where $N_{body}$ and $N_{act}$ denote the numbers of body-level and action-level categories, respectively. This task requires the model to distinguish highly similar subtle motions under diverse recording conditions.

\textbf{\emph{Evaluation Metrics.}} We adopt evaluation metrics widely used in action recognition to systematically assess model performance at different label levels. Specifically, we use accuracy and the F1 score as primary metrics. In particular, we report the accuracy of Top-$1$ (Acc@1) and Top-$5$ (Acc@5) at both the body level and the action level, together with macro-F1 and micro-F1 scores, following previous work on video classification~\cite{nie2017enhancing,guo2024benchmarking}. Since our dataset exhibits class imbalance, we also report mean class accuracy (MCA) following~\cite{yan2018spatial,duan2022revisiting}. MCA computes the accuracy for each class independently and then averages the results, thus treating all classes equally regardless of their sample sizes.

\begin{figure*}[h]
\centering

\includegraphics[width=0.9\linewidth]{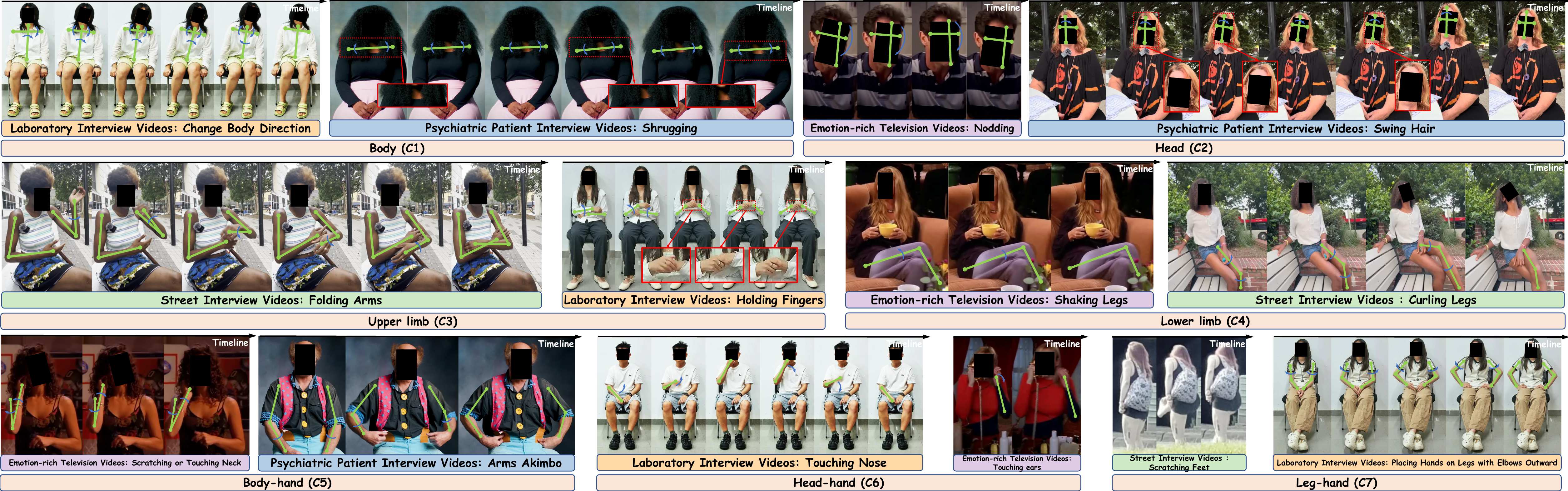}

\caption{Example video clips and annotations for the MMA-82-Rec dataset.}
\label{fig:MAR_visual}
\vspace{-0.5cm}
\end{figure*}

\textbf{\emph{MMA-82-Rec Dataset.}} We select a total of 39,816 video clips from the four different sources of MMA-82, namely laboratory interview videos, psychiatric patient interview videos, street interview videos, and emotion-rich television videos, containing 27,509, 7,011, 3,925, and 1,371 video clips, respectively. Together, these clips constitute the recognition benchmark MMA-82-Rec. Fig.~\ref{fig:distribution_MAR} illustrates the dataset statistics from three perspectives. The left panel shows the distribution of body-level action instances, where \emph{head} and \emph{upper limb} account for more than two-thirds of the total. This is consistent with human behavioral patterns, as nonverbal information is often conveyed primarily through head and upper-limb movements. The middle panel presents the frame-length distribution of action instances in each body-level category, revealing subtle differences in temporal duration across categories. The right panel shows the action-level category distribution, which exhibits a pronounced long-tail pattern. Such imbalance increases the difficulty of learning robust representations and may lead to biased model performance. We further provide representative examples of MMA-82-Rec in Figure~\ref{fig:MAR_visual}. The dataset covers a wide variety of videos with diverse sources, viewpoints, and backgrounds. In the visualization, green lines highlight the moving joints, while blue arrows indicate motion trends. As can be observed, micro-actions usually occur within very small spatial ranges, further underscoring the difficulty of the task. 

For benchmarking, MMA-82-Rec is split into training, validation, and test sets with a ratio of 3:1:1, while preserving a similar action distribution across the three splits. To prevent identity leakage, \emph{each subject is assigned exclusively to a single split}. Detailed dataset statistics and split information are reported in Table~\ref{tab:data_split_count}. To comprehensively evaluate recognition performance under both seen and unseen conditions, we further define two evaluation settings: \textbf{In-Domain} and \textbf{Cross-Domain}.

\begin{table}[]
\centering
\caption{Statistics of the MMA-82-Rec data splits.}
\vspace{-0.2cm}
\resizebox{0.85\linewidth}{!}{
\setlength{\tabcolsep}{2.5pt}
\begin{tabular}{l|l||llc|l}
\thickhline
\rowcolor{mygray}
Data Sources                     & Splits      & Clips & Duration & Avg. Length & \#Subj. \\ \hline\hline
\multirow{4}{*}{Lab Interview} & Train   & 15,820 & 11.85h   & 2.70s     & 152     \\
                            & Val & 5,636  & 4.12h    & 2.63s     & 52      \\
                            & Test    & 6,053  & 4.52h    & 2.69s     & 25      \\  
                            & Total      & 27,509 & 20.48h   & 2.68s     & 229     \\ \hline 
\multirow{4}{*}{Psychiatric Interview} & Train   & 4,203  & 3.19h   & 2.73s     & 8        \\
                             & Val & 1,398  & 1.13h    & 2.92s     & 6        \\
                             & Test    & 1,410  & 1.40h    & 3.56s     & 5        \\  
                             & Total      & 7,011  & 5.72h    & 2.94s     & 19        \\ \hline 
\multirow{4}{*}{Street Interview}       & Train   & 2,358  & 1.18h    & 1.80s     & 12        \\
                            & Val & 792    & 0.44h    & 2.00s     & 7        \\
                            & Test    & 775    & 0.47h    & 2.20s     & 7        \\  
                            & Total      & 3,925  & 2.10h    & 1.92s     & 26        \\ \hline 
\multirow{4}{*}{Emotion videos} & Train   & 795 & 0.37h   & 1.70s     & 52        \\
                                & Val & 247 & 0.11h   & 1.58s     & 79        \\
                                & Test    & 329 & 0.15h   & 1.68s     & 49       \\  
                                & Total      & 1,371 & 0.64h & 1.67s     & 180       \\ \hline 
\rowcolor[HTML]{f8f9fa}
\multicolumn{2}{c||}{Total} & 39,816 & 28.94h   & 2.62s     & 454        \\ \thickhline
\end{tabular}
}
\label{tab:data_split_count}
\vspace{-0.7cm}
\end{table}

\textbf{\textit{1) In-Domain}}

This setting evaluates the model's basic recognition ability when the training and test samples are drawn from the same domain and thus share similar scene characteristics. As baselines, we first adopt PoseConv3D (PoseC3D)~\cite{duan2022pyskl}, a widely used framework for skeleton-based action recognition. PoseC3D consists of two main stages: 2D skeleton sequences are first converted into a 3D heatmap volume, and then a 3D CNN is used to extract spatio-temporal features for action classification. We further include a pure RGB-based video recognition model, GC-TSM~\cite{hao2022group}, which enhances TSM~\cite{lin2019tsm} with a family of efficient element-wise calibrators. More details about the baselines and training protocols are provided in the Supplementary. Table~\ref{tab:mr_in_result} reports the benchmark results of these two methods under the in-domain setting.

Several observations can be drawn from the results. First, on the overall MMA-82-Rec benchmark, the RGB-based GC-TSM achieves better action-level recognition performance than the skeleton-based PoseC3D, improving the test Top-1 Acc from 56.62 to 60.43, and MCA from 36.77 to 38.98. This indicates that appearance and contextual cues are helpful for recognizing subtle micro-actions. Second, the performance varies substantially across different sub-datasets. The laboratory interview subset is the easiest one, where both methods achieve their best results. In contrast, performance drops markedly on the psychiatric and street interview subsets, and becomes particularly poor on the emotion video subset. These results suggest that domain complexity, viewpoint variation, background clutter, and more spontaneous motion patterns introduce substantial challenges for micro-action recognition. 

\begin{table*}[b!]
\centering
\vspace{-0.5cm}
\caption{Baseline Results for the Micro-Action Recognition under the in-domain setting.}
\vspace{-0.2cm}
\setlength\tabcolsep{3pt}
\resizebox{1.0\linewidth}{!}{
\begin{tabular}{l||l||cc|cc|cc|cc|cc||cc|cc|cc|cc|cc}
\thickhline
\rowcolor{mygray}
& & \multicolumn{10}{c||}{\textbf{Action Level}} & \multicolumn{10}{c}{\textbf{Body Level}}   \\ \cline{3-22}
\rowcolor{mygray}
\multicolumn{1}{c||}{\multirow{-1}{*}{Sub-Dataset}} &
\multicolumn{1}{c||}{\multirow{-1}{*}{Method}}
                             & \multicolumn{2}{c|}{Top-$1$ Acc}
                             & \multicolumn{2}{c|}{Top-$5$ Acc}
                             & \multicolumn{2}{c|}{MCA}
                             & \multicolumn{2}{c|}{Macro F1}
                             & \multicolumn{2}{c||}{Micro F1}
                               
                             & \multicolumn{2}{c|}{Top-$1$ Acc}
                             & \multicolumn{2}{c|}{Top-$5$ Acc}
                             & \multicolumn{2}{c|}{MCA}
                             & \multicolumn{2}{c|}{Macro F1}
                             & \multicolumn{2}{c}{Micro F1}    \\ \cline{3-22}
\rowcolor{mygray}
& & Val & Test & Val & Test & Val & Test & Val & Test & Val & Test & Val & Test & Val & Test & Val & Test & Val & Test & Val & Test  \\ \thickhline
                               
\multirow{2}{*}{MMA-82-Rec (All)} & Skeleton & 54.43 & 56.62 & 80.44 & 80.45 & 36.26 & 36.77 & 37.35 & 39.39 & 54.43 & 56.62 & 79.04 & 81.93 & 99.18 & 99.11 & 69.62 & 71.71 & 71.12 & 74.62 & 79.04 & 81.93 \\  
 & RGB & 57.59 & 60.43 & 83.01 & 86.14 & 37.39 & 38.98 & 35.64 & 39.56 & 57.59 & 60.43  & 79.18 & 83.17 & 98.79 & 98.95 & 67.24 & 69.48 & 67.77 & 71.98 & 79.18 & 83.17 \\ \hline\hline 
\multirow{2}{*}{Laboratory Interviews} & Skeleton & 57.81 & 64.84 & 84.28 & 87.81 & 39.82 & 44.88 & 41.77 & 47.03 & 57.81 & 64.84 & 81.74 & 86.73 & 99.59 & 99.80 & 75.19 & 79.78 & 76.16 & 81.87 & 81.74 & 86.73 \\
 & RGB & 62.01 & 68.15 & 87.31 & 92.83 & 43.00 & 48.01 & 39.69 & 46.48 & 62.01 & 68.15 & 83.61 & 86.75 & 99.13 & 99.45 & 73.59 & 75.96 & 73.85 & 77.66 & 83.61 & 86.75 \\ \hline
\multirow{2}{*}{Psychiatric Interviews} & Skeleton & 50.64 & 41.63 & 75.46 & 68.79 & 27.33 & 15.74 & 24.93 & 17.65 & 50.64 & 41.63 & 76.97 & 77.73 & 98.71 & 99.01 & 57.29 & 47.46 & 52.54 & 51.33 & 76.97 & 77.73 \\
 & RGB & 52.00 & 46.74 & 75.89 & 72.13 & 24.77 & 14.10 & 17.33 & 13.64 & 52.00 & 46.74 & 74.25 & 81.99 & 98.00 & 99.08 & 52.64 & 44.26 & 45.34 & 45.36 & 74.25 & 81.99 \\ \hline
\multirow{2}{*}{Street Interviews} & Skeleton & 44.82 & 40.77 & 68.18 & 70.71 & 19.55 & 19.49 & 21.20 & 20.70 & 44.82 & 40.77 & 69.95 & 71.35 & 97.60 & 98.84 & 45.44 & 43.39 & 49.09 & 49.52 & 69.95 & 71.35 \\
 & RGB & 42.80 & 39.87 & 69.70 & 73.55 & 13.79 & 12.75 & 12.09 & 12.56 & 42.80 & 39.87 & 63.26 & 68.90 & 97.85 & 97.29 & 37.53 & 37.02 & 36.25 & 38.23 & 63.26 & 68.90 \\ \hline
\multirow{2}{*}{Emotion Videos} & Skeleton & 29.55 & 7.29 & 60.32 & 17.93 & 19.25 & 3.72 & 19.01 & 3.11 & 29.55 & 7.29 & 58.30 & 35.87 & 97.57 & 88.45 & 38.50 & 19.26 & 39.74 & 17.89 & 58.30 & 35.87 \\
 & RGB & 35.63 & 25.53 & 67.61 & 52.89 & 21.83 & 12.20 & 22.19 & 11.05 & 35.63 & 25.53 & 57.09 & 55.93 & 98.38 & 93.01 & 31.04 & 33.20 & 29.69 & 31.01 & 57.09 & 55.93 \\ 
\thickhline
\end{tabular}
}
\label{tab:mr_in_result}
\end{table*}

\begin{table*}[t!]
\centering
\caption{Baseline Results for the Micro-Action Recognition under the cross-domain settings (zero-shot and few-shot).}
\vspace{-0.2cm}
\setlength\tabcolsep{2pt}
\resizebox{0.85\linewidth}{!}{
\begin{tabular}{l||l||ccccc|ccccc}
\thickhline
\rowcolor{mygray}
 & & \multicolumn{5}{c|}{\textbf{Action Level}} & \multicolumn{5}{c}{\textbf{Body Level}}   \\ \cline{3-12}
\rowcolor{mygray}
\multicolumn{1}{c||}{\multirow{-2}{*}{Source}} & \multicolumn{1}{c||}{\multirow{-2}{*}{Task}}
                             & Top-$1$ Acc    
                             & Top-$5$ Acc
                             & MCA
                             & Macro F1
                             & Micro F1 
                               
                             & Top-$1$ Acc     
                             & Top-$5$ Acc
                             & MCA
                             & Macro F1
                             & Micro F1       
                             \\ \thickhline
\multirow{4}{*}{Psychiatric Interviews} 
& \cellcolor[HTML]{f8f9fa}Zero-Shot 
& \cellcolor[HTML]{f8f9fa}27.30 
& \cellcolor[HTML]{f8f9fa}50.28
& \cellcolor[HTML]{f8f9fa}14.25
& \cellcolor[HTML]{f8f9fa}7.96
& \cellcolor[HTML]{f8f9fa}27.30
& \cellcolor[HTML]{f8f9fa}68.58
& \cellcolor[HTML]{f8f9fa}96.74 
& \cellcolor[HTML]{f8f9fa}49.40
& \cellcolor[HTML]{f8f9fa}42.43
& \cellcolor[HTML]{f8f9fa}68.58
\\
& 1-Shot     & 30.27 & 58.62 & 22.82 & 13.31 & 30.28 & 72.48 & 93.22 & 55.50 & 44.47 & 72.48      \\ 
& 5-Shot    & 30.12 & 58.60 & 22.72 & 13.29 & 30.12 & 72.60 & 93.28 & 55.43 & 44.41 & 72.60     \\ 
& 10-Shot     & 30.13 & 58.59 & 22.71 & 13.24 & 30.12 & 72.64 & 93.30 & 55.42 & 44.48 & 72.64    \\ \hline\hline 

\multirow{4}{*}{Street Interviews} 
& \cellcolor[HTML]{f8f9fa}Zero-Shot 
& \cellcolor[HTML]{f8f9fa}20.65 
& \cellcolor[HTML]{f8f9fa}44.90 
& \cellcolor[HTML]{f8f9fa}10.65 
& \cellcolor[HTML]{f8f9fa}6.91 
& \cellcolor[HTML]{f8f9fa}20.65 
& \cellcolor[HTML]{f8f9fa}53.16
& \cellcolor[HTML]{f8f9fa}92.65 
& \cellcolor[HTML]{f8f9fa}38.68 
& \cellcolor[HTML]{f8f9fa}35.12 
& \cellcolor[HTML]{f8f9fa}53.16 
\\
& 1-Shot  & 21.38 & 45.64 & 15.60 & 12.40 & 21.38 & 54.95 & 84.46 & 38.92 & 37.27 & 54.95      \\ 
& 5-Shot & 21.58 & 46.10 & 16.17 & 13.28 & 21.58 & 64.91 & 95.61 & 40.33 & 17.00 & 64.91     \\ 
& 10-Shot  & 22.52 & 45.92 & 17.76 & 14.23 & 22.52 & 65.64 & 97.22 & 39.37 & 38.91 & 65.64      \\ \hline\hline 

\multirow{4}{*}{Emotion Videos} 
& \cellcolor[HTML]{f8f9fa}Zero-Shot 
& \cellcolor[HTML]{f8f9fa}14.13
& \cellcolor[HTML]{f8f9fa}32.98
& \cellcolor[HTML]{f8f9fa}6.63
& \cellcolor[HTML]{f8f9fa}4.18
& \cellcolor[HTML]{f8f9fa}14.13
& \cellcolor[HTML]{f8f9fa}43.92
& \cellcolor[HTML]{f8f9fa}90.43
& \cellcolor[HTML]{f8f9fa}25.18
& \cellcolor[HTML]{f8f9fa}23.77
& \cellcolor[HTML]{f8f9fa}43.92
\\
& 1-Shot  & 17.26 & 39.77 & 14.00 & 11.62 & 17.26 & 41.43 & 79.84 & 27.11 & 25.13 & 41.43      \\ 
& 5-Shot & 17.48 & 41.75 & 15.20 & 12.10 & 17.49 & 42.35 & 90.80 & 26.93 & 24.56 & 42.35      \\ 
& 10-Shot  & 17.70 & 42.53 & 15.43 & 12.26 & 17.70 & 42.03 & 92.30 & 26.95 & 24.39 & 42.03      \\ \hline\hline 

\end{tabular}
}
\label{tab:mr_cro_result}
\end{table*}

\textbf{\textit{2) Cross-Domain}}

To further evaluate model robustness and transferability, we introduce a more challenging Cross-Domain setting, in which training and validation are conducted in one domain while testing is performed in a different domain. In our benchmark, the laboratory interview videos are used for training and validation, whereas the other three domains are separately used as test sets. Under this setting, we further define two evaluation protocols: \textbf{Zero-Shot} and \textbf{Few-Shot}. We adopt PoseC3D as the baseline model.

In the \textbf{Zero-Shot} setting, the model is trained only on the laboratory interview data, without accessing any samples from the target test domains. That is, the PoseC3D model trained on the laboratory interview videos is directly applied to predict the micro-action category of each video clip in the other three domains. Since the target domains are entirely unseen during training, this setting evaluates the model's direct cross-domain generalization capability. In the \textbf{Few-Shot} setting, the model is allowed to access a small number of labeled samples from the target domain for adaptation. Specifically, we follow a standard $K$-shot protocol with $K \in \{1,5,10\}$, where $K$ labeled clips are randomly sampled from each class in the target domain for adaptation, and the remaining clips are used for evaluation. To reduce the effect of random sampling, each experiment is repeated over 20 independent runs, and the average performance is reported.

Table~\ref{tab:mr_cro_result} reports the results under the cross-domain setting. In particular, under the zero-shot protocol, we observe a substantial performance drop compared with the in-domain setting, indicating that direct transfer across domains remains highly challenging for micro-action recognition. Although performance improves in the few-shot setting, even with only one labeled sample per class, the gains gradually diminish as $K$ increases from 1 to 5 and 10. This suggests that simply increasing the number of labeled target-domain samples provides only limited improvement for micro-action understanding in uncontrolled scenarios. These observations underline the need for more robust generalization and adaptation strategies for realistic micro-action recognition.

\vspace{-0.4cm}
\subsection{Multi-label Micro-Action Detection} \label{sec:MAD}

\textbf{\textit{Task Definition.}} 
Multi-label Micro-Action Detection (MAD) aims to detect and classify all micro-action instances in an untrimmed video. Formally, given an untrimmed video $\mathbf{V}_{utr}$ with a total number of $T$ frames, the ground-truth action annotations are represented as
\vspace{-0.5em}
\begin{equation}
    \Phi_g = \left\{ \phi_i = (t_{s}^{i}, t_{e}^{i}, c_i) \right\}_{i=1}^{N_g},
    \vspace{-0.5em}
\end{equation}
where $t_{s}^{i}$ and $t_{e}^{i}$ denote the start and end times of the $i$-th micro-action instance, $c_i \in \mathcal{C}$ denotes its category label, $\mathcal{C}$ is the set of all micro-action categories, and $N_g$ is the total number of ground-truth instances. The goal of MAD is to predict a set of action proposals
\vspace{-0.5em}
\begin{equation}
    \Phi_p = \left\{ \hat{\phi}_i = (\hat{t}_{s}^{i}, \hat{t}_{e}^{i}, \hat{c}_i, s_i) \right\}_{i=1}^{N_p},
    \vspace{-0.5em}
\end{equation}
where $\hat{t}_{s}^{i}$ and $\hat{t}_{e}^{i}$ denote the predicted temporal boundaries, $\hat{c}_i$ denotes the predicted category label, $s_i$ denotes the confidence score, and $N_p$ is the number of predicted action instances.

\begin{table}[h]
\centering
\caption{MMA-82-Det Dataset Statistics. ``Duration'', ``Avg. Video'', ``Avg. Insta. Du'' and ``Avg. Insta.'' denote the total video duration, average video duration,  average instance duration, and average number of instances per video, respectively.}
\vspace{-0.2cm}
\setlength\tabcolsep{2pt}
\resizebox{1.0\linewidth}{!}{
\begin{tabular}{c||cccccc}
\thickhline
\rowcolor{mygray}
Split & Videos & Instances & Duration & Avg. Video & Avg. Insta. Du. & Avg. Insta. \\ \thickhline
Training & 7,839 & 28,240 & 33.05h & 15.18s & 3.62s & 3.60 \\
Validation & 2,127 & 7,236 & 8.92h & 15.10s & 3.86s & 3.40 \\
Testing & 1,214 & 4,282 & 4.97h & 14.73s & 3.59s & 3.53 \\ \hline
\rowcolor[HTML]{f8f9fa}
All & 11,180 & 39,758 & 46.93h & 15.11s & 3.66s & 3.56 \\ \thickhline

\end{tabular}
}
\label{tab:mad_data_count}
\vspace{-0.5cm}
\end{table}

\begin{figure*}[t]
\centering
\vspace{-1em}
\includegraphics[width=0.9\linewidth]{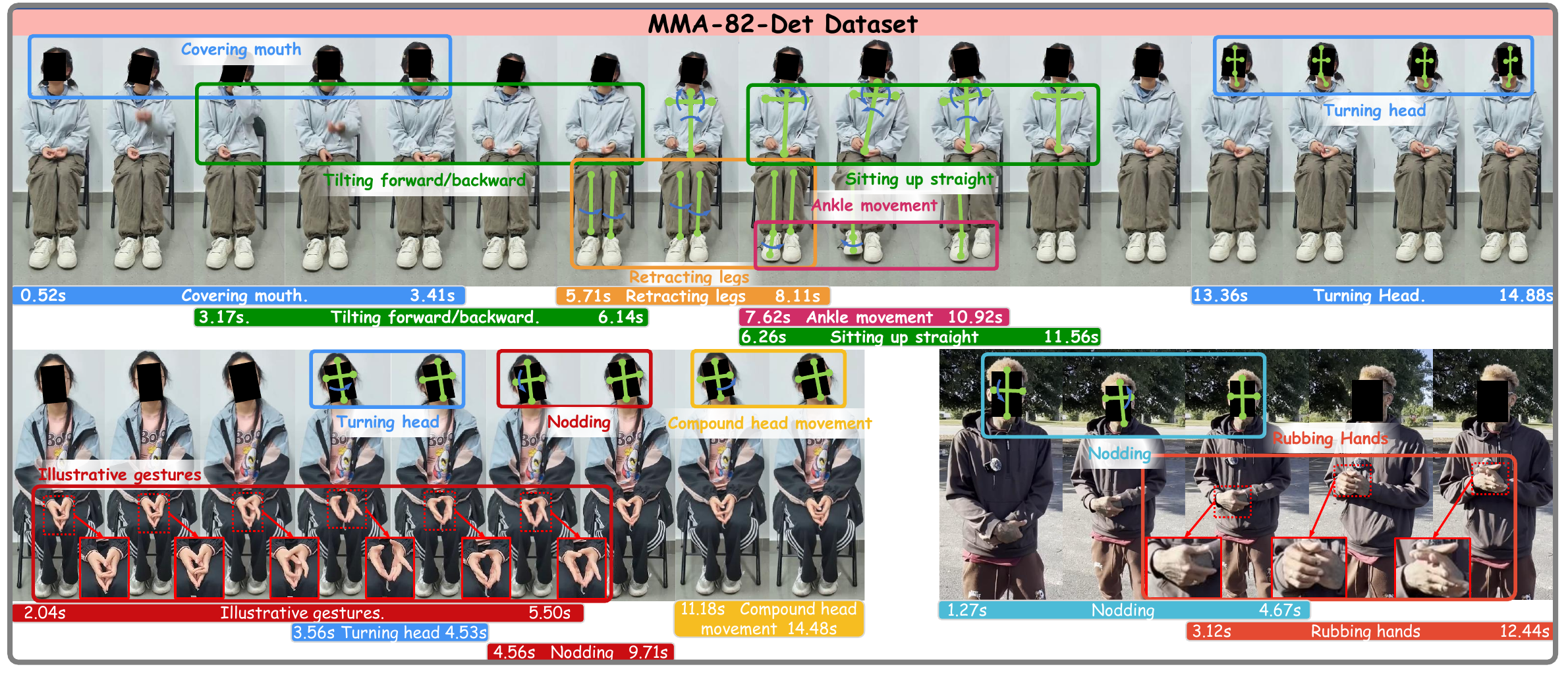}
\vspace{-0.3cm}
\caption{Example video clips and annotations from the MMA-82-Det dataset.}
\label{fig:mad_instance}
\vspace{-0.5cm}
\end{figure*}

\textbf{\textit{MMA-82-Det Dataset.}} 
We construct the MMA-82-Det dataset by cropping 11,180 videos from the original MMA-82 collection, with video durations ranging from 0.6 seconds to 1.5 minutes and a total of 39,758 action instances. Fig.~\ref{fig:distribution_MAD} presents the statistical analysis of MMA-82-Det, while Table~\ref{tab:mad_data_count} summarizes its detailed split information. Similar to MMA-82-Rec, MMA-82-Det also exhibits a noticeably imbalanced distribution over different body-level and action-level categories, with a pronounced long-tail pattern at the fine-grained action level. Such imbalance increases the difficulty of learning robust temporal representations and may lead to biased detection performance toward frequent categories.  In addition, the temporal lengths of action instances vary considerably, and multiple micro-actions often co-occur or appear in rapid succession within a single video, making precise temporal localization particularly challenging. On average, each video contains 3.56 micro-action instances, and each instance lasts approximately 3.66 seconds. We also visualize several representative detection examples in Fig.~\ref{fig:mad_instance}, which illustrate the diversity of recording conditions, action transitions, and co-occurring micro-actions in both controlled and real-world scenarios. These properties make MMA-82-Det a challenging benchmark for multi-label micro-action detection, requiring models to jointly address fine-grained category discrimination and accurate temporal localization.

\textbf{\textit{Evaluation Metrics.}} We adopt Detection mean Average Precision (Detection-mAP)~\cite{tan2022pointtad,li2025mmad} as the evaluation metric. Detection-mAP primarily measures the match between predicted action instances and ground-truth in localization and category recognition. It evaluates the model's performance by calculating the temporal intersection over union (tIoU) between predicted and ground-truth segments. Specifically, we compute Detection-mAP across multiple tIoU thresholds and report the mAP for tIoU from 0.1 to 0.9 (in steps of 0.1). Considering the two-level micro-actions, we evaluate at both the body-level and action-level and report the average mAP.

\textbf{\textit{Baselines.}} We use AdaTAD~\cite{liu2024end} as the baseline for MMA-82-Det. AdaTAD is a state-of-the-art end-to-end framework for temporal action detection that addresses the memory and computational constraints typically associated with scaling up TAD models. Unlike traditional feature-based methods that rely on pre-extracted backbone features, AdaTAD integrates a lightweight adapter directly into the frozen backbone. These adapters allow the framework to capture complex temporal dependencies across long video sequences while maintaining efficient memory usage.

\begin{table}[]
\centering
\caption{The experimental results on the MMA-82-Det dataset.}
\vspace{-0.2cm}
\setlength\tabcolsep{2pt}
\resizebox{1.0\linewidth}{!}{
\begin{tabular}{c||cccc|cccc||c}

\thickhline
\rowcolor{mygray}
 & \multicolumn{4}{c|}{\texttt{Action-level}} & \multicolumn{4}{c||}{\texttt{Body-level}} &  \\ \cline{2-9}
\rowcolor{mygray}
 \multirow{-2}{*}{Backbone} & @0.2 & @0.5 & @0.7 & Avg & @0.2 & @0.5 & @0.7 & Avg & \multirow{-2}{*}{\texttt{AVG}}\\ \thickhline

VideoMAE-S & 20.88 & 12.72 & 5.56 & 12.09 & 48.18 & 28.78 & 13.91 & 25.44 & 18.77 \\
VideoMAE-B & 22.62 & 14.67 & 6.32 & 13.59 & 50.95 & 30.46 & 12.23 & 29.13 & 21.36 \\
VideoMAE-L & 22.74 & 15.60 & 7.68 & 14.98 & 55.48 & 33.01 & 14.06 & 31.83 & 23.41 \\
\rowcolor[HTML]{f8f9fa}
VideoMAE-H & 26.53 & 17.56 & 7.55 & 16.08 & 54.71 & 33.64 & 14.17 & 30.05 & 23.07 \\ 
\thickhline

\end{tabular}}
\label{tab:detection_results}
\vspace{-0.5cm}
\end{table}

\textbf{\textit{Results.}} Table~\ref{tab:detection_results} reports a comparison of our performance using AdaTAD~\cite{liu2024end} on the MMA-82-Det dataset with VideoMAE~\cite{tong2022videomae} backbones of different scales. As the model scales up from VideoMAE-S to VideoMAE-L, the average mAP (AVG) continues to increase, rising from 18.77 to 23.41. This indicates that larger pre-trained models can provide more discriminative spatio-temporal representations for micro-action detection. Notably, compared to VideoMAE-L, VideoMAE-H shows only a small improvement at the action level and a slight drop at the body level, which may be due to increased optimization difficulty resulting from the larger model size.

\vspace{-0.2cm}
\section{Emotion with Micro-Actions}
Video-based emotion recognition has been widely studied in affective computing and human-centered AI. Existing methods mainly rely on facial affective cues, such as facial expressions and micro-expressions, together with holistic video context, including body posture, scene context, and temporal dynamics. However, the role of \emph{micro-actions} remains largely unexplored. Compared with micro-expressions, micro-actions capture subtle whole-body-level movements that may reveal spontaneous reactions and fine-grained affective changes, and can provide complementary behavioral evidence when facial cues are occluded or insufficient. This section addresses \textbf{RQ3} and \textbf{RQ4} by examining whether micro-actions are associated with emotional states and whether they provide complementary affective cues beyond facial micro-expressions. To answer them, we conduct empirical analysis on the emotion-rich subset of MMA-82 and evaluate the contribution of micro-actions to video emotion recognition.

{\bf \emph{Emotion-Action Dataset.}} We build the Emotion-Action dataset based on CAER~\cite{lee2019context}, which originally contains seven emotion categories: \emph{anger}, \emph{disgust}, \emph{fear}, \emph{happy}, \emph{neutral}, \emph{sad}, and \emph{surprise}. We extend it in two aspects. First, we introduce an additional category, \emph{melancholy}, to better capture subtle differences within negative emotions. In our definition, \emph{melancholy} denotes a milder and more implicit affective state than \emph{sad}. Second, we annotate the micro-actions in each video clip to explicitly link emotional states with subtle body movements. We select 100 clips for each category, resulting in a total of 800 clips.

\begin{figure}[h]
  \centering
    \includegraphics[width=0.9\linewidth]{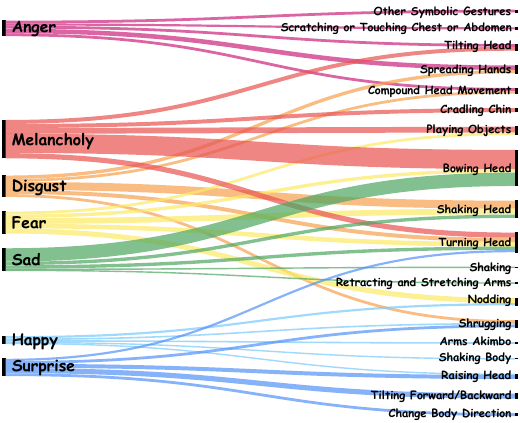}
    \vspace{-0.2cm}
    \caption{Sankey-style visualization of the Top-5 micro-actions associated with each emotion category. Flow width denotes relative importance.}
    \label{fig:importance}\
    \vspace{-0.4cm}
  \end{figure}

{\bf \emph{Exploration of Emotion-Action Relationships.}} To answer \textbf{RQ3}, we analyze the relationship between emotion categories and micro-actions using decision-tree-based feature importance and Sankey-style visualization. More details for the decision-tree construction are provided in the Appendix. Based on the results of the decision-tree,  Fig.~\ref{fig:importance} visualizes the Top-5 micro-actions for each emotion. Two observations can be made. First, each emotion is strongly associated with several specific micro-actions, indicating that subtle whole-body movements provide meaningful cues for affective understanding. Second, similar emotions share overlapping micro-action patterns. For example, \emph{sad} and \emph{melancholy} both correlate strongly with \emph{bowing head} and \emph{turning head}, suggesting partially common behavioral patterns. At the same time, their differences are also informative: \emph{sad} is more associated with \emph{shaking} and \emph{retracting and stretching arms}, whereas \emph{melancholy} is more related to \emph{playing objects} and \emph{cradling chin}. This suggests that \emph{sad} tends to involve more explicit negative bodily expressions, while \emph{melancholy} is reflected through milder and more inward subtle movements.

We further validate these findings from an interdisciplinary perspective using \emph{Laban Movement Analysis} (LMA)~\cite{laban1971mastery}\footnote{It is a classical framework that connects human movement with psychological and emotional states.}. The high-importance micro-actions identified above show strong consistency with established movement psychology theory~\cite{shafir2016emotion,melzer2019we}. For instance, positive emotions are often associated with upward, open, and rhythmic movements, which align well with actions such as \emph{Nodding}, \emph{Tilting Forward/Backward}, \emph{Raising Head}, and \emph{Shrugging}\footnote{ In our analysis, \emph{Nodding} and \emph{Tilting Forward/Backward} are closely related to positive affect and are well aligned with the LMA element of \emph{Rhythmicity}. Similarly, \emph{Raising Head} and \emph{Shrugging} correspond to the \emph{Up and Rise} element, while \emph{Arms Akimbo} is related to the \emph{Spread} element.}. In contrast, negative emotions are more related to downward or contractive movements (LMA descriptor of \emph{Head-Drop}), such as \emph{Bowing Head}. This agreement suggests that micro-actions are not only statistically correlated with emotions, but also psychologically meaningful.

\begin{table}[]
\centering
\caption{Performance comparison of facial micro-expression and micro-action based methods.}
\vspace{-0.2cm}
\begin{tabular}{l|c||cc}
\thickhline
\rowcolor{mygray}
Task & Method & Top-1 Acc & F1 \\
\thickhline
Micro-Expression Only & DeepFace~\cite{serengil2026boosted} & 22.86 & 17.54 \\
Micro-Action Only & TSM~\cite{lin2019tsm} & 32.38 & 31.86 \\ \hline
\rowcolor[HTML]{f8f9fa}
Both & DeepFace + TSM & \textbf{32.86} & \textbf{32.36} \\
\thickhline
\end{tabular}
\label{tab:emo_results}
\vspace{-0.5cm}
\end{table}


{\bf \emph{Micro-Actions for Emotion Recognition.}} To answer \textbf{RQ4}, we compare facial micro-expression cues, micro-action cues, and their combination for video emotion recognition. Specifically, we evaluate three settings: \emph{(1) Micro-Expression Only}, \emph{(2) Micro-Action Only}, and \emph{(3) Both}. The first uses only facial micro-expression cues as the baseline, the second uses only micro-action cues, and the third combines both to assess whether micro-actions provide complementary affective information beyond micro-expressions. Table~\ref{tab:emo_results} shows that micro-actions alone already achieve meaningful emotion recognition performance, indicating a strong relationship between subtle body movements and emotional states. More importantly, combining micro-actions with micro-expressions further improves recognition over the facial baseline, demonstrating that micro-actions provide complementary behavioral evidence for affective analysis. It is worth noting that the micro-action cues used in this experiment are derived from ground-truth annotations rather than model predictions. This design is intended to verify the intrinsic value of micro-actions for emotion recognition and to examine whether they provide useful affective information when accurately identified.

\vspace{-0.2cm}
\section{Conclusion}

In this paper, we have presented \textbf{MMA-82}, a comprehensive multi-domain benchmark that extends MA-52 toward more realistic micro-action analysis. MMA-82 expands the taxonomy from 52 to 82 fine-grained categories and covers four distinct domains beyond controlled laboratory settings. Based on this benchmark, we established two core tasks, \textbf{MMA-82-Rec} and \textbf{MMA-82-Det}, together with in-domain, cross-domain, few-shot, and zero-shot evaluation protocols. Extensive experiments show that current methods still face significant challenges under domain shift, long-tailed distributions, and temporal localization. We further showed that micro-actions are strongly associated with emotional states and provide complementary cues to facial micro-expressions for emotion recognition. We hope MMA-82 will serve as a useful foundation for future research on subtle action understanding, temporal action detection, cross-domain generalization, and multimodal affective behavior analysis.

\section{Ethical Considerations}
MMA-82 is constructed for academic research on micro-action understanding. For the laboratory interview data, informed consent was obtained from the recruited participants. For videos collected from publicly available online sources, including psychiatric interviews, street interviews, and emotion-rich television videos, we followed the corresponding platform policies and used them only for research purposes. We carefully screened the data and excluded clearly identifiable minors, inappropriate content, and samples unsuitable for academic use. Our annotations focus only on observable micro-action categories and temporal boundaries, without collecting private personal information or assigning identity-related, clinical, or psychological labels. For data release, we provide annotations and source references when raw videos are restricted by copyright or platform policies.

{\small
\bibliographystyle{IEEEtran}

\bibliography{main}
}

\newpage
\clearpage

\appendices

\section{Baseline Implementation Details}

In this appendix, we provide the implementation details of all baselines evaluated on the MMA-82 benchmark.

\subsection{Implementation Details of PoseConv3D}

We use PoseConv3D~\cite{duan2022pyskl} as the baseline for micro-action recognition based on skeleton. It uses a 3D convolutional neural network to process skeleton heatmap sequences. The model backbone is based on ResNet3D, with 17 input channels. It consists of three stages, containing 4, 6, and 3 layers respectively, and uses dilated convolutions in the second and third stages to expand the temporal receptive field. The spatial domain is downsampled with a stride of 2. The final features are fed into a 512-channel I3D~\cite{carreira2017quo} classification head, which outputs classification results for 82 classes. In the data pipeline, 32 frames are sampled uniformly from the raw skeleton sequence at a resolution of 64×64. The training process uses the SGD~\cite{alistarh2017qsgd} optimizer with an initial learning rate of 0.2, a weight decay of 3e-4, and 60 training epochs.

\subsection{Implementation Details of GC-TSM}

For RGB input, we use GC-TSM~\cite{hao2022group} as the baseline. GC-TSM embeds a lightweight group contextualization module into the TSM backbone network, enabling it to capture more spatio-temporal information critical to recognizing subtle actions while maintaining high computational efficiency. Specifically, we use ResNet-50 as the backbone and select 8 frames from each video as input. Additionally, SGD is used as the optimizer, with an initial learning rate of 0.01 and a weight decay of 1e-4. Decay is applied at the 20th and 40th epochs, with a total of 50 training epochs and a batch size of 16.

\subsection{Implementation Details of AdaTAD}

For the micro-action detection task, we use the end-to-end action detection framework AdaTAD on the MMA-82-Det dataset. Its backbone uses a vision transformer with an adapter, with VideoMAE pre-trained on Kinetics-400 serving as the base weights. The model uses sliding windows for sampling, with a window size setting of 256. After center-cropping, the data is resized to 160×160 and fed into the network. AdaTAD uses AdamW for optimization, with the initial learning rates for both the backbone and the adapter set to 1e-4, and the model trained for 20 epochs. During the inference stage, Soft-NMS is used for post-processing, with the NMS threshold parameter set to 0.5. It retains the top 200 highest-confidence bounding boxes and uses a voting strategy to further refine the predicted temporal boundaries.

\section{More Results Analysis on MAD}

\begin{figure}[h]
\centering

\subfigure[False Negative Analysis]{
    \includegraphics[width=\linewidth]{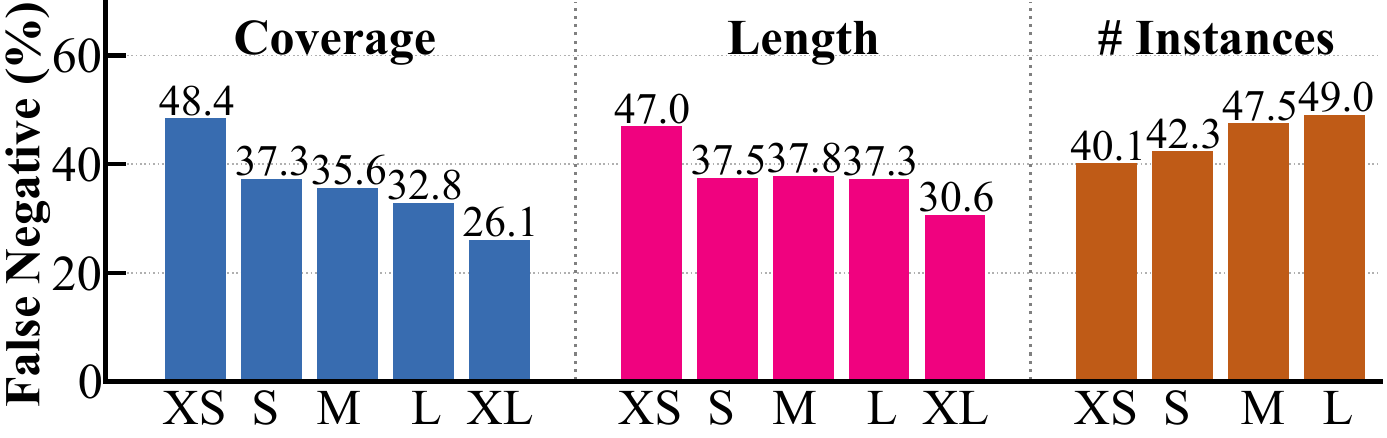}
    \label{fig:false_negative_analysis}
}
\hfill
\subfigure[False Positive Analysis]{
    \includegraphics[width=\linewidth]{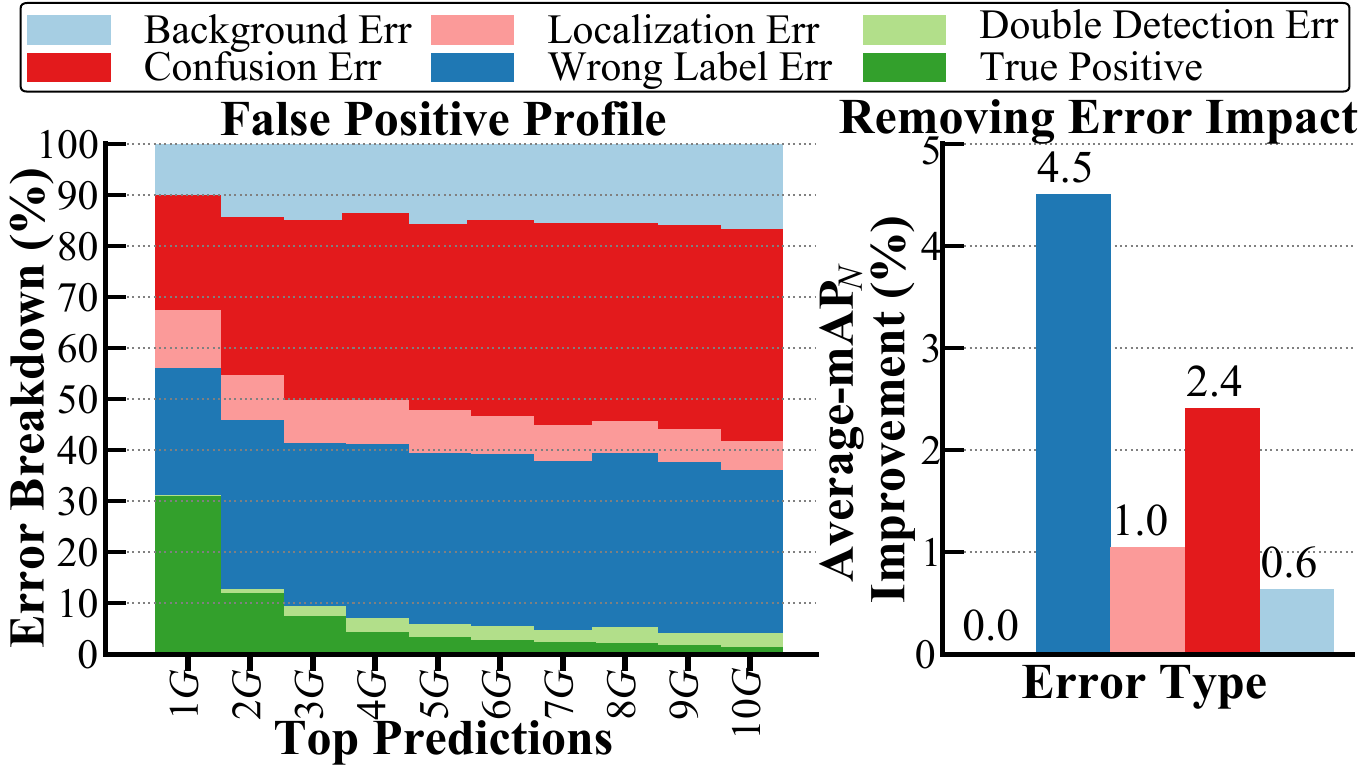}
    \label{fig:false_positive_analysis}
}
\hfill
\subfigure[Sensitivity Analysis]{
    \includegraphics[width=\linewidth]{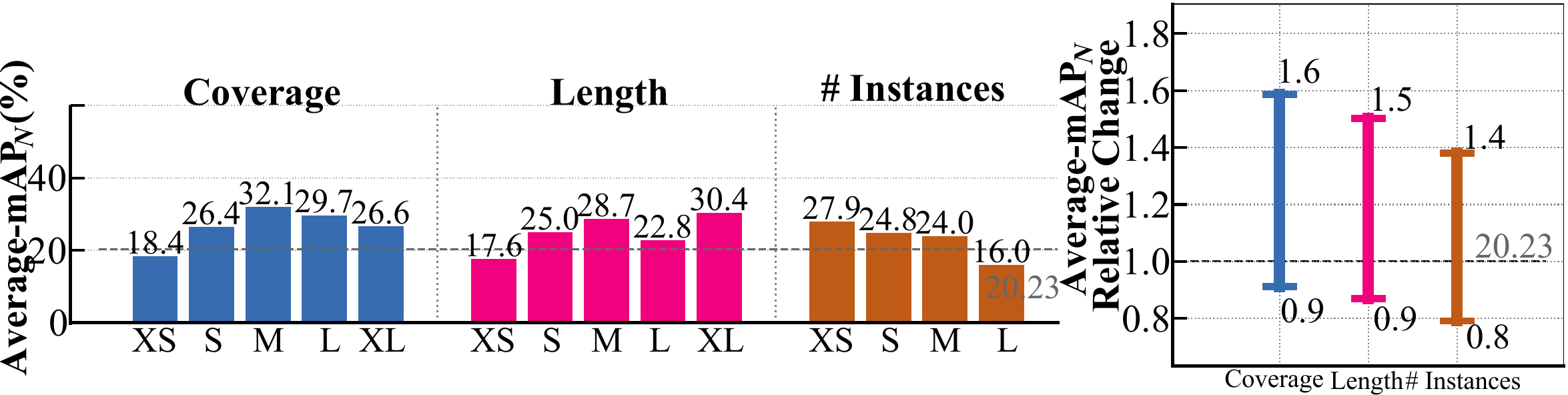}
    \label{fig:sensitivity_analysis}
}

\caption{Error analysis for the MAD task.}
\label{fig:error_analysis}
\end{figure}

Following TAD conventions~\cite{liu2024end, zhang2022actionformer} and micro-action detection task specifications~\cite{li2025mmad}, we used detection analysis tools~\cite{alwassel2018diagnosing} to evaluate the model's performance on VideoMAE-H at the action level. 

\textbf{1) False Negative Profiling:} As shown in Figure~\ref{fig:false_negative_analysis}, we report the model’s false negatives (FN) across three dimensions: coverage, length, and instances. The baseline achieves a low FN in most dimensions. As coverage and length increase, the FN decreases; however, as the number of instances increases, the FN rises. 

\textbf{2) False Positive Profiling:} As shown in Figure~\ref{fig:false_positive_analysis}, in the left figure, we analyze the false positives at tIoU=0.5. These errors fall into five major categories, with ``Confusion Error'' and ``Wrong Label Error'' accounting for the majority. Furthermore, in the right figure, we break down the impact of each error type on the average mAP. Removing ``Wrong Label Error'' brings the most significant performance improvement (+4.5\%), followed by ``Confusion Error'' (+2.4\%). Although action localization is crucial, distinguishing between similar actions and filtering out background noise are also critical tasks that cannot be overlooked. 

\textbf{3) Sensitivity Profiling:} To evaluate the model’s robustness, we analyze the performance sensitivity across different scales, as shown in Figure~\ref{fig:sensitivity_analysis}. The model performs stably under varying coverage levels, with performance maximizing at 32.1\% when coverage is set to M. Performance tends to improve as action duration increases, but decreases as action density increases, which is consistent with real-world patterns.

\section{More Analysis for Emotion with Micro-Actions}

\subsection{Examples of Dataset}

\begin{figure*}[t]
\centering

\includegraphics[width=\linewidth]{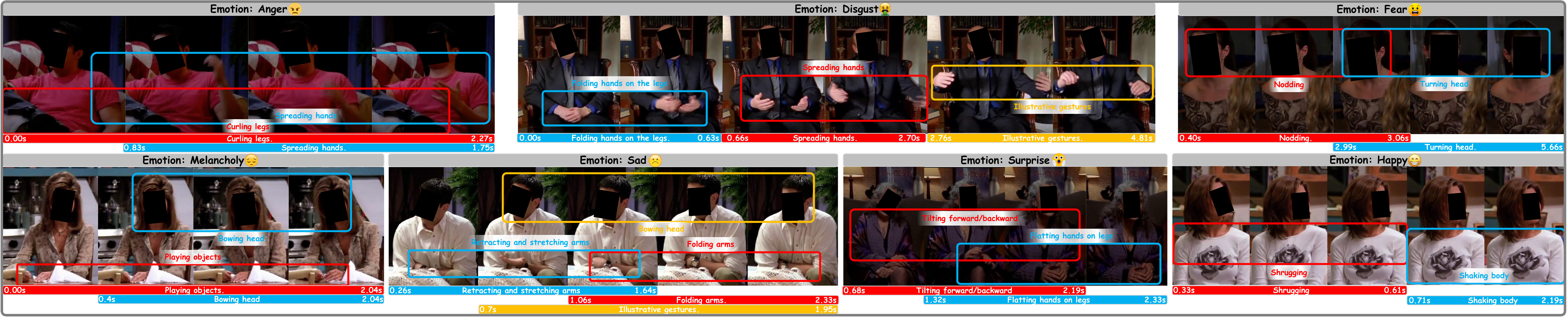}

\caption{Video clip examples from the Emotion-rich television video collection.}
\label{fig:emotion_instance}
\end{figure*}

As shown in Figure~\ref{fig:emotion_instance}, we visualize some example videos from the Emotion-Action Dataset. Each video generally contains one or more action instances, which together reflect non-spontaneous human emotions. For example, when a person is sad, they tend to show sinking and contracting movements such as ``Retracting and stretching arms'' and ``Folding arms,'' which is consistent with LMA theory~\cite{laban1971mastery}. When a person is happy, they exhibit rising, light-weight, rhythmic movements such as ``Shrugging'' and ``Shaking body.''

\subsection{Details of Micro-Actions for Emotion Recognition}

To verify the contribution of micro-actions to emotion recognition, we use DeepFace and TSM to recognize emotions in facial expressions and micro-actions, respectively. The results are reported in Table~\ref{tab:emo_results}. Specifically, we first use DeepFace to recognize emotions in cropped facial video frames and retain the logit vectors $l_{face}$ after removing the final Softmax layer. Similarly, for micro-action video frames, TSM predicts emotions by capturing the spatiotemporal interactions of micro-actions, and we similarly retain the logit vectors $l_{mas}$. Subsequently, rather than concatenating features directly in high dimensions, we adopt a late fusion strategy, which mitigates issues of feature dimensional mismatch and redundant noise. We concatenate these two independent logits to obtain a concatenated logits $l_{fusion}$. This vector is then fed into a lightweight MLP for fine-tuning, resulting in an optimized model.

\begin{figure*}[t]
\centering

\includegraphics[width=\linewidth]{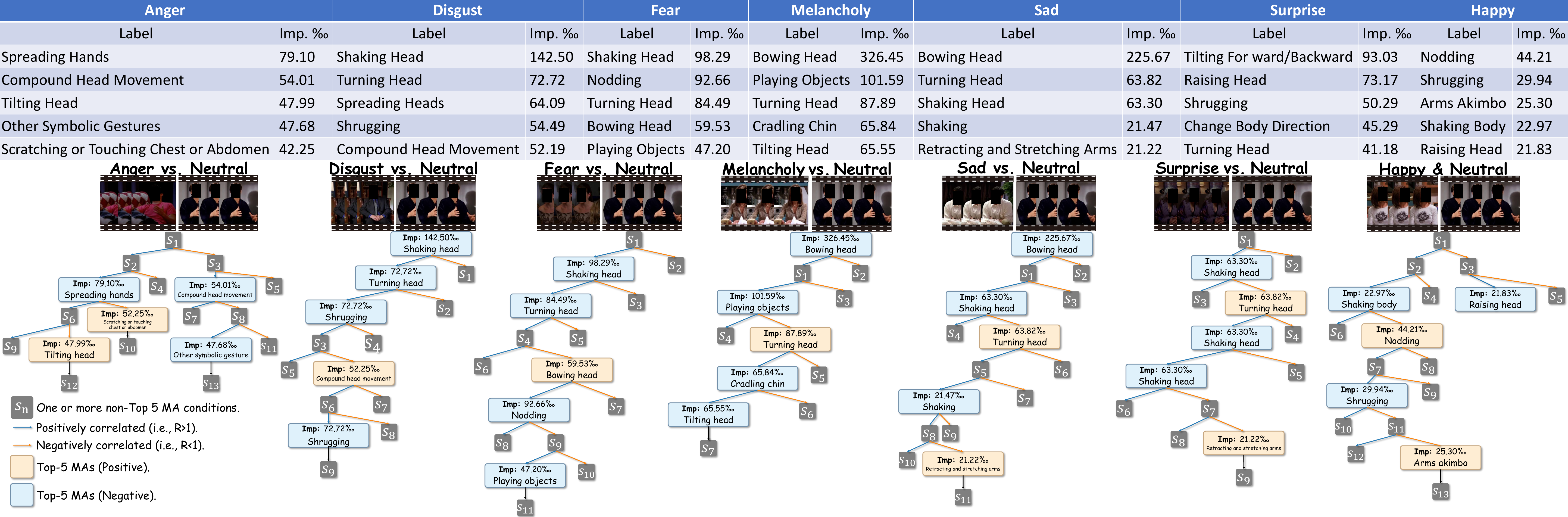}

\caption{Decision-tree-based visualization of emotion discrimination and the Top-5 micro-actions most strongly associated with each emotion category. For each emotion, the figure shows the decision process under the One-Versus-Neutral (OVN) setting and highlights the five micro-actions with the highest importance scores.}
\label{fig:emotion_tree}
\end{figure*}

\subsection{Details of Emotion-Action Tree}

Due to the wide variation in the frequency of natural micro-actions among subjects, using absolute action counts directly would result in emotion-related key actions being overrepresented among high-frequency micro-actions. Therefore, we treat each video clip as a document and each action as a word, and process the data using Term Frequency–Inverse Document Frequency (TF-IDF)~\cite{aizawa2003information}.
First, for each video containing multiple annotated micro-actions, we first compute a TF-IDF~\cite{aizawa2003information} feature vector and use it as the input to the decision tree. To better distinguish subtle and visually similar emotions (\ie, \emph{melancholy} and \emph{sad}), we adopt a \emph{One-Versus-Neutral} (OVN) strategy instead of the conventional \emph{One-Versus-Rest} (OVR) setting. This design provides a more stable and interpretable way to analyze the relationship between emotions and micro-actions.

For each OVN tree, we use the CART to construct it, with the choice of splitting points based on Gini impurity. To measure the global contribution of micro-action features within the tree, we calculate the feature importance for each micro-action, which is determined by the weighted sum of the Gini impurity values obtained at the nodes where the micro-action feature serves as a splitting node. We use Figure~\ref{fig:emotion_tree} to present the decision process of the decision tree for emotion discrimination, while highlighting the Top-5 micro-actions most strongly associated with each emotion category. 

Based on the learned trees, we compute the importance of each micro-action for distinguishing a target emotion. Since feature importance reflects only contribution magnitude rather than association direction, we further define the following conditional probability ratio:
\begin{equation}
    R = \frac{P(MA \mid E_t)}{P(MA \mid \overline{E_t})},
\end{equation}
where $E_t$ denotes the target emotion and $\overline{E_t}$ denotes all non-target emotions. When $R \geq 1$, the micro-action $MA$ is regarded as positively correlated with the target emotion.

To further verify whether the micro-actions selected by the OVN decision tree described above retain strong emotional representational capabilities, we use these filtered features as input to an MLP for emotion classification. The pseudocode for the entire algorithm is shown in Algorithm~\ref{alg:mlp}.

\begin{algorithm}[h]
\caption{MLP-based Emotion Recognition}
\label{alg:mlp}
\begin{algorithmic}[1]

\Require TF-IDF feature matrix $X \in \mathbb{R}^{n \times m}$
\Require Emotion labels $y \in \{1, \ldots, 7\}$
\Require Selected MA set $M$

\Ensure Predicted labels $\hat{y}$

\State Normalize MA frequency features
\State Construct TF-IDF representation $X$
\State Split $X$ and $y$ into training, validation, and test sets
\State Initialize MLP parameters $\theta$

\For{$epoch = 1$ to $E$}
    \For{each mini-batch $(x_b, y_b)$}
        \State $h_b \gets \mathrm{ReLU}(W_1x_b + b_1)$
        \State $h_b \gets \mathrm{Dropout}(h_b)$
        \State $z_b \gets W_2h_b + b_2$
        \State $L \gets \mathrm{CrossEntropy}(z_b, y_b)$
        \State Update $\theta$ using Adam optimizer
    \EndFor

    \State Evaluate validation performance
    \State Update best model if F1 score improves
\EndFor

\State $\hat{y} \gets$ Predict test labels using the best model
\State \Return $\hat{y}$

\end{algorithmic}
\end{algorithm}

We conduct three sets of experiments to validate its effectiveness. First, we input all TF-IDF values as features into the MLP. Second, we retain the Top-5 MAs, set the others to 0, recalculate the TF-IDF, and input it into the MLP. Finally, we set all Top-5 MAs to 0, recalculate the TF-IDF, and input it into the MLP. We report the validation results in Table~\ref{tab:ovn_results}.

\begin{table}[h]
\centering
\caption{Results of Top-5 Micro-Action and Emotion}
\resizebox{1.0\linewidth}{!}{
\begin{tabular}{ll||ccc}
\thickhline
\rowcolor{mygray}
No. & Experience & Acc & Delta & F1 \\ \thickhline
1 & Base Results      & 0.271 & 0 & 0.277 \\
2 & no Top-5 MAs only & 0.186 & -0.086 & 0.147 \\ \hline
\rowcolor[HTML]{f8f9fa}
3 & Top-5 MAs only    & \textbf{0.379} & \textbf{+0.107} & \textbf{0.350} \\ \thickhline
\end{tabular}
}
\label{tab:ovn_results}
\end{table}

This indicates that using only non-Top-5 MAs significantly reduces the accuracy of emotion recognition, whereas considering only the Top-5 MAs significantly improves it. This demonstrates that the actions we identified are strongly correlated with the corresponding emotions and can effectively enhance the performance of emotion recognition.

\vfill

\end{document}